\documentclass[10pt,twocolumn,letterpaper]{article}

\usepackage{cvpr}
\usepackage{times}
\usepackage{epsfig}
\usepackage{graphicx}
\usepackage{amsmath}
\usepackage{amssymb}

\usepackage{url}

\cvprfinalcopy 

\def\cvprPaperID{****} 

\begin{document}

\title{Local Feature Detectors, Descriptors, and Image Representations: A Survey}

\author{Yusuke Uchida\\
The University of Tokyo\\
Tokyo, Japan\\
}

\maketitle

\begin{abstract}
With the advances in both stable interest region detectors and robust and distinctive descriptors, local feature-based image or object retrieval has become a popular research topic.
The other key technology for image retrieval systems is image representation such as the bag-of-visual words (BoVW), Fisher vector, or Vector of Locally Aggregated Descriptors (VLAD) framework.
In this paper, we review local features and image representations for image retrieval.
Because many and many methods are proposed in this area, these methods are grouped into several classes and summarized.
In addition, recent deep learning-based approaches for image retrieval are briefly reviewed.
\end{abstract}


\section{Introduction}

Image retrieval is the problem of searching for digital images in large databases.
It can be classified into two types: text-based image retrieval and content-based image retrieval~\cite{rui_jvcir99}.
Text-based image retrieval (or concept-based image retrieval) refers to an image retrieval framework, where the images first are annotated manually, and then text-based Database Management Systems (DBMS) is utilized to perform retrieval~\cite{tam_pr84}.
However, the rapid increase of the size of image collection in the early 90's brought two difficulties.
One is that the vast amount of labor is required in manually annotating the images.
The other difficulty is the subjectivity of human perception;
it sometimes happens that different people perceive the same image differently, resulting in different annotation results or different query keywords in search.
This makes text-based image retrieval results less effective.

\subsection{From Text-based to Content-based Image Retrieval}
In order to overcome these difficulties, Content-Based Image Retrieval (CBIR)~\cite{gud_com95} or Query By Image Content (QBIC)~\cite{fli_com95} was proposed.
In CBIR, images are automatically annotated with their own visual content by feature extraction process.
The visual content includes colors~\cite{pas_wacv96, hua_cvpr97, kas01}, shapes~\cite{bel_tpami02}, textures~\cite{won_etri02}, or any other information that can be derived from the image itself.
Extracted features representing visual content are indexed by high multi-dimensional indexing techniques to realize large-scale image retrieval~\cite{rui_jvcir99}.

\begin{figure}[tb]
	\centering
	\includegraphics[width=0.8\linewidth]{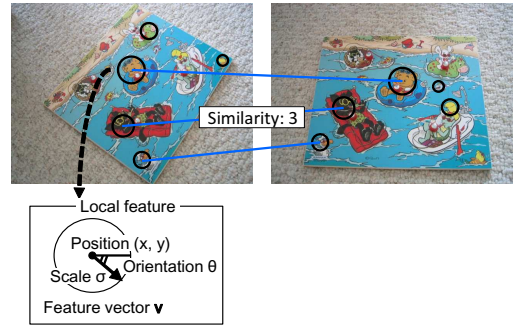}
	\caption{An example of local feature-based image retrieval.}
	\label{fig:example}
\end{figure}

\subsection{Local Feature-based Image Retrieval}
Although lots of image features are proposed in the middle of 90s in order to improve CBIR system, most of these features are \textit{global} and therefore have difficulty in dealing with partial visibility and extraneous features.
In order to handle partial visibility and transformations such as image rotation and scaling, a pioneer work on local feature-based image retrieval was done in~\cite{sch_tpami97}.
In this framework, interest points are autoatically detected from an image, and then feature vectors are computed at the interest points.
In search step, each of feature vectors extracted from a query image votes scores to matched referece features whose feature vectors are similar to the query feature vector.
In local feature-based image retrieval, many local features are used in search, it is very robust against partial occulusion.

Figure~\ref{fig:example} shows a toy example of local feature-based image retrieval, where the similarity of the two images is to be calculated.
Fisrtly, local features are extracted from both images.
Then, these local features are \textit{matched} to generate pairs of local features whose feautre vectors are similar.
In Figure~\ref{fig:example}, there are three pairs after matching.
The most simple way to define the similarity between these two images is to use the number of the matched pairs: three in this case.

All of the local feature-based image retrieval system involves two important processes: local feature extraction and image representation.
In local feature extraction, certain local features are extracted from an image.
And then, in image representation, these local features are integrated or aggregated into a vector representation in order to calculate similarity between images\footnote{Images are not necessarily represented by a single vector. Some methods simply defines the similarity between two sets of local features instead of explicitly integrating them into vectors.}.

\begin{figure*}[tb]
	\centering
	\includegraphics[width=\linewidth]{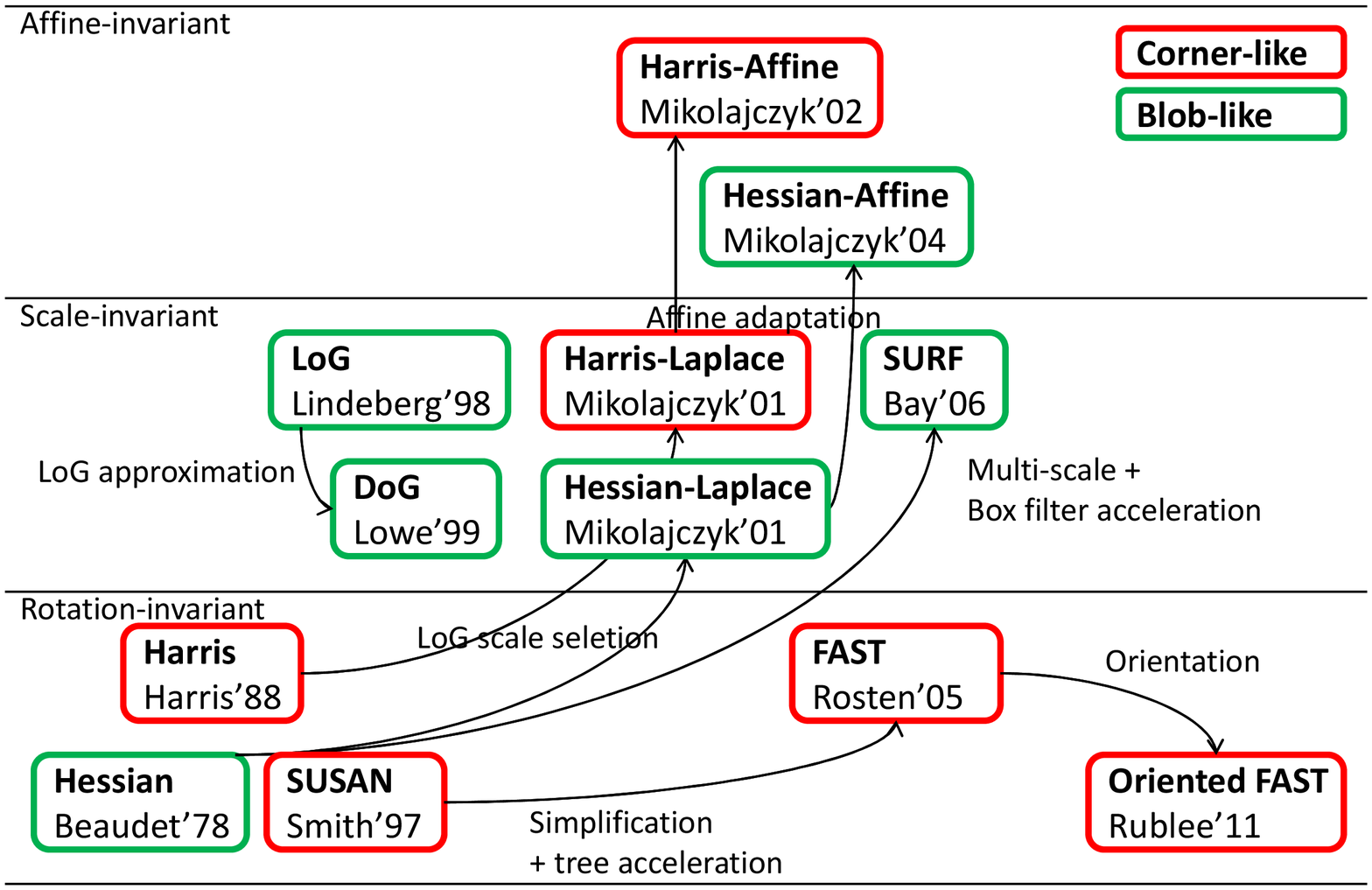} \\
	(a) Feature detectors. \\
	\includegraphics[width=\linewidth]{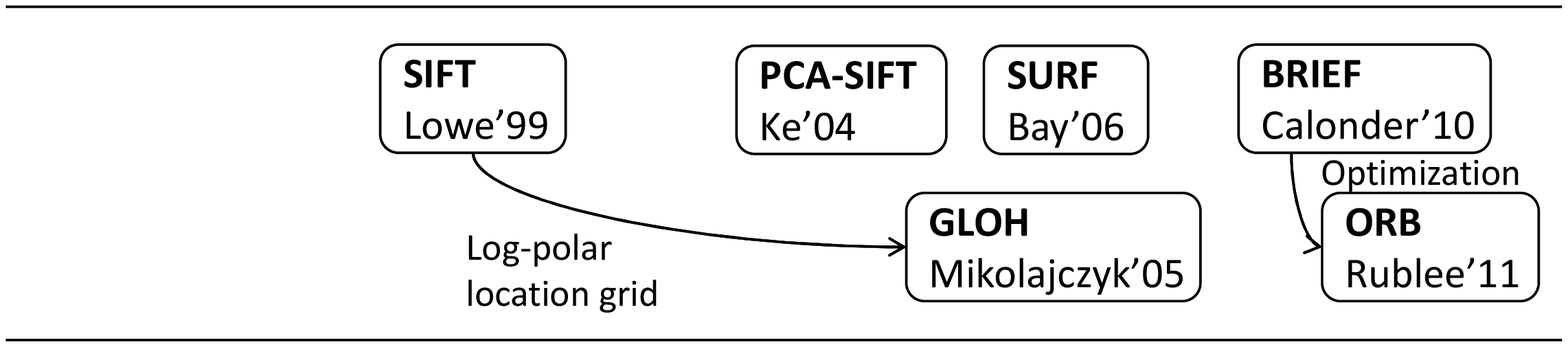} \\
	(b) Feature descriptors. \\
	\includegraphics[width=\linewidth]{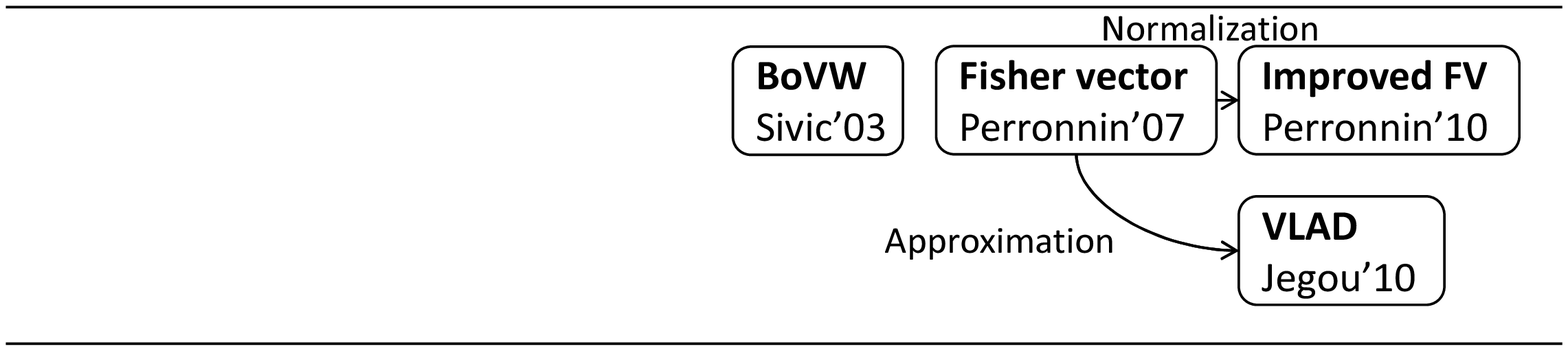} \\
	(c) Image representations. \\
	\caption{The overview of the history of local features and image representations. Only an especially important part of local features and image representations is shown.}
	\label{fig:overlook}
\end{figure*}

In this paper, we review local features and image representations for image retrieval.
In Section~\ref{sec:local_features}, various local features are introduced, which are the basis of recent local feature-based retrieval or recognition frameworks.
In Section~\ref{sec:image_representation}, various image representations are explained.
In Figure~\ref{fig:overlook}, the overview of the history of local features and image representations.

\section{Local Features}
\label{sec:local_features}
In this section, local features used in local feature-based image retrieval are reviewed.
Local features are characterized by the combination of \textit{feature detector} and \textit{feature descriptor}.
A feature detector finds feature points/locations, e.g. $(x, y)$, or feature regions, e.g. $(x, y, \sigma)$, where $\sigma$ denotes the scale of the region.
A feature descriptor extracts multi-dimensional feature vectors from the detected points or regions.
While feature detectors and feature descriptors can be used in arbitrary combinations, specific combinations are usually used such as the DoG detector and the SIFT descriptor, or multi-scale FAST detector and the BRIEF descriptor.
In order to make local features invariant to rotation, the orientation of a local feature is estimated in many local features.
In this paper, the algorithms for the orientation estimation are included in the feature descriptor part, not in the detector part.
While we focus on the systematic summary of local features, there are complementary comparative evaluations of local features \cite{mik_pami05, mik_ijcv05, gau_ijcv11, mik_eccv12, hei_eccv12, bia_icce13, cha_mmsp13, can_dsp13, fil_visapp14, bia_dsp15}.
These evaluations are very usuful to grasp the performance of local features.

\subsection{Feature Detectors}
Feature detectors find multiple feature points or feature regions from an image.
Feature detectors can be characterized by two factors: region type and invariance type.
The region type represents the shape of a detected point or region such as corner or blob.
The invariance type here represents to which transformations the detector is robust.
The transformation can be a rotation, a similarity transformation, or an affine transformation.
It is important to choose a feature detector with a specific invariance suitable for the problem we are solving.

\subsubsection{Harris, Harris-Laplace, and Harris-Affine Detector}
\textbf{Harris detector}~\cite{har_avc88} is one of the most famous corner detectors, which extends Moravec's corner detector.
The original idea of the Moravec detector is to detect a pixel such that there is no nearby similar patch to the patch centered on the pixel.
We assume a grayscale image $I$ as an input.
Let $I(u, v)$ denote the intensity of the pixel at $(u, v)$ in $I$.
Following the idea of Muravec detector, let $E(x, y)$ denote the weighted sum of squared differences caused by a shift $(x, y)$:
\begin{equation}
\label{eq:harris_e}
E(x, y) = \sum_{u, v} w(u, v) \left( I(u+x, v+y) - I(u, v) \right)^2,
\end{equation}
where $w(u, v)$ is a window function.
The term $I(u+x, v+y)$ can be approximated by a Taylor expansion as
\begin{equation}
I(u+x, v+y) \approx I(u, v) + I_x(u, v) x + I_y(u, v) y,
\end{equation}
where $I_x$ and $I_y$ denotes the partial derivatives of $I$ with respect to $x$ and $y$.
Using this approximation, Eq.~(\ref{eq:harris_e}) can be written as
\begin{equation}
E(x, y) \approx \sum_{u, v} w(u, v) \left( I_x(u, v) x + I_y(u, v) y \right)^2.
\end{equation}
This can be re-written in matrix form:
\begin{equation}
E(x, y) \approx (x, y) M (x, y)^{\top},
\end{equation}
where
\begin{equation}
M = \sum_{u, v} w(u, v)
\begin{bmatrix}
I_x^2 & I_x I_y \\
I_x I_y & I_y^2 \\
\end{bmatrix}
\end{equation}
If $E(x, y)$ becomes large in any shift $(x, y)$, it indicates a corner.
This can be judged using the eigenvalues of $M$.
Letting $\alpha$ and $\beta$ denote the eigenvalues of $M$, $E(x, y)$ increases in all shift if both $\alpha$ and $\beta$ are large.
Instead of evaluating $\alpha$ and $\beta$ directly, it is proposed to use $\mathrm{Tr}(M) = \alpha + \beta$ and $\mathrm{Det}(M) = \alpha \beta$ for efficiency~\cite{har_avc88};
the corner response $R$ is defined as:
\begin{equation}
R = \mathrm{Det}(M) - k \left( \mathrm{Tr}(M) \right)^2.
\end{equation}
Thresholding on the values of $R$ and performing non-maxima supporession, the Harris corners are detected from the input image.

The Harris detector is effective in the situation where scale change does not occur like tracking or stereo matching.
However, as the Harris detector is very sensitive to changes in image scale, it is not appropriate for image retrieval, where the sizes of objects in query and reference images are frequently different.
Therefore scale-invariant deature detector is essential for robust image recognition or retrieval.

\textbf{Harris-Laplace detector}~\cite{mik_iccv01} is a scale-adapted Harris detector.
It firstly detects candidate feature points using the Harris detector on multiple scales (multi-scale Harris detector).
Then, these candidate feature points are verified using the Laplacian to check whether the detected scale is maxima or not in the scale direction (cf. LoG Detector).
The Harris-Laplace detector detects corner-like structures.

\textbf{Harris-Affine detector}~\cite{mik_eccv02, mik_ijcv04} is a affine-invariant feature detecctor.
It firstly detects feature points using the Harris-Laplace detector.
Then, iteratively refine these regions to affine regions using the second moment matrix as proposed in \cite{lin_iccv95, lin_ivc97}.
The resulting Harris-Affine regions are characterized by ellipses.

\subsubsection{Hessian, Hessian-Laplace, and Hessian-Affine Detector}
\textbf{Hessian detector} \cite{bea_ijcpr78} searches for image locations that have strong derivatives in two orthogonal directions.
It is based on the matrix of second derivatives, namely Hessian:
\begin{equation}
H(x, y, \sigma) = 
\begin{bmatrix}
L_{xx}(x, y, \sigma)	& L_{xy}(x, y, \sigma) \\
L_{xy}(x, y, \sigma) & L_{yy}(x, y, \sigma) \\
\end{bmatrix},
\end{equation}
where $L(x, y, \sigma)$ is an image smoothed by a Gaussian kernel $G(x, y, \sigma)$:
\begin{equation}
L(x, y, \sigma) = G(x, y, \sigma) * I(x, y).
\end{equation}
The Hessian detector detects $(x, y)$ as feature point such that the determinant of the Hessian $H$ is local-maxima comapred with neighboring 8 pixels:
\begin{equation}
\det(H) = L_{xx} L_{yy} - L_{xy}^2.
\end{equation}

\textbf{Hessian-Laplace detector}~\cite{mik_iccv01} is a scale-adapted Hessian detector.
It firstly detects candidate feature points using the Hessian detector on multiple scales (multi-scale Hessian detector).
Then, these candidate feature points are selected according to the Laplacian in the same way as Harris-Laplace.
Note that the trade of the Hessian matrix is identical the Laplacian:
\begin{equation}
\mathrm{tr}(H) = L_{xx} + L_{yy}.
\end{equation}
The Hessian-Laplace detector detects blob-like structures similar to the LoG or DoG detectors explained later.
It is claimed that these methods often detect feature points on edges while the Hessian-Laplace does not, owing to the use of the determinant of the Hessian~\cite{mik_ijcv05}.

\textbf{Hessian-Affine detector}~\cite{mik_ijcv05} is a affine-invariant feature detecctor and is imilar in spirt as the Harris-Affine detector.
It firstly detects feature points using the Hessian-Laplace detector.
Then, iteratively refine these regions to affine regions using the second moment matrix as done in the Harris-Affine detector.

\subsubsection{LoG Detector}
Detecting scale-invariant regions can be accomplished by searching for stable regions across all possible scales, using a continuous function of scale known as scale space.
A scale space representation is defined by
\begin{equation}
L(x, y, \sigma) = G(x, y, \sigma) * I(x, y),
\end{equation}
where $G(x, y, \sigma)$ is a Gaussian kernel:
\begin{equation}
G(x, y, \sigma) = \frac{1}{2 \pi \sigma^2} \exp \left( -(x^2 + y^2)/ 2\sigma^2 \right).
\end{equation}
In \cite{lin_ijcv98}, a blob detector that searches for scale space extrema of a scale-normalized Laplacian-of-Gaussian (LoG) $\sigma^2 \nabla^2 L$, where
\begin{equation}
\nabla^2 L = L_{xx} + L_{yy}.
\end{equation}
The term $\sigma^2$ is the normalization term, which normalizes response of LoG filter among different scales.


\subsubsection{DoG Detector}
Scale-Invariant Feature Transform (SIFT)~\cite{low_cvpr99, low04}\footnote{The SIFT algorithm includes both of detection and detection. In this paper, they are distinguished by using the terms \textit{the SIFT detector} and \textit{the SIFT descriptor}.} is one of the most widely used local features due to its robustness.
In detection of SIFT, it is proposed to use scale-space extrema in the Difference-of-Gaussian (DoG) function instead of LoG in order to efficiently detect stable keypoint.
The DoG $D(x, y, \sigma)$ can be computed from the difference of two nearby scales separated by a constant multiplicative factor $k$:
\begin{align}
D(x, y, \sigma)	&= \left( G(x, y, k \sigma) - G(x, y, \sigma) \right) * I(x, y) \\
				&= L(x, y, k \sigma) - L(x, y, \sigma).
\end{align}
The DoG is a close approximation to the scale-normalized LoG $\sigma^2 \nabla^2 L$, which is shown using the heat diffusion equation:
\begin{equation}
\frac{\partial L}{\partial \sigma} = \sigma \nabla^2 L.
\end{equation}
The term ${\partial L}/{\partial \sigma}$ can be approximated using the difference of nearby scales at $k \sigma$ and $\sigma$:
\begin{equation}
\frac{\partial L}{\partial \sigma} \approx \frac{L(x, y, k \sigma) - L(x, y, \sigma)}{k \sigma - \sigma}.
\end{equation}
Thus, we get:
\begin{equation}
L(x, y, k \sigma) - L(x, y, \sigma) \approx (k - 1) \sigma^2 \nabla^2 L.
\end{equation}.
The above equation shows that the response of DoG is already scale-normalized.
Thus, DoG detector detects $(x, y, \sigma)$ is a feature region if the response of $D(x, y, \sigma)$ is a local maxima or minima by comparing its 26 neighbors in terms of $x$, $y$, and $\sigma$ dimensions.
In \cite{low04}, the DoG is efficiently calculated using image pyramid.

The detected region $(x, y, \sigma)$ is further refined to sub-pixel and sub-scale accuracy by fitting a 3D quadratic to the scale-space Laplacian~\cite{bro_bmvc02, low04}.
After this refinement, detected regions are filterd out according to absolute values of their DoG responses and cornerness measures similar to the Harris detector in order to remove low contrast or edge regions~\cite{low04}.

\subsubsection{SURF Detector}
Speeded Up Robust Features (SURF)~\cite{bay_eccv06, bay_cviu08} or fast Hessian detector is efficient approximation of the Hessian-Laplace detector.
In \cite{bay_eccv06, bay_cviu08}, it is proposed to approximate with box filters the Gaussian second-order partial derivatives $L_{xx}$, $L_{xy}$, and $L_{yy}$, which are required in the calculation of the determinant of the Hessian.
These box filters can be efficiently calculated using \textit{integral images} \cite{vio_cvpr01}.
The SURF detector detects a scale-invariant blob-like features similar to the Hessian-Laplace detector.
While the Hessian-Laplace detector uses the determinant of Hessian to select the location of the features and uses LoG to determine the characteristic scale, the SURF detecter uses the determinant of Hessian for both similar to the DoG detector.


\subsubsection{FAST Detector}
Most of the local binary features employ fast feature detectors.
The Features from Accelerated Segment Test (FAST)~\cite{ros_iccv05, ros_eccv06, ros_pami10} detector is one of such extremely efficient feature detectors.
It can be considered as a simplied version of the Smallest Uni-value Segment Assimilating Nucleus Test (SUSAN) detector~\cite{smi_ijcv97}, which detects pixels such that there is few similar pixels around the pixels.
The FAST Detector detects pixels that are brighter or darker than neighboring pixels based on the accelerated segment test as follows.
For each pixel $p$, the intensities of 16 pixels on a Bresenham circle of radius 3 are compared with that of $p$, and are classified into three tyeps: \textit{brighter}, \textit{similar}, and \textit{darker}.
If there is least $S$ connected pixels on the circle which are classified to brighter or darker, $p$ is detected as a corner.
In order to avoid detecting edges, $S$ must be larger than nine and the FAST with $S = 9$ (FAST-9) is usually used.

In \cite{ros_iccv05}, it is proposed to accelerate this test by firstly checking the four pixels at the top, bottom, left, and right on the circle, achieving early rejection of the test.
In \cite{ros_eccv06}, the segment test is further sped up by using a decision tree.
By using a decision tree, the test is optimized to reject candidate pixels very quickly, realizing extremely fast feature detection.
In \cite{rub_iccv11}, it is proposed to filter out the detected FAST features according to their Harris scores.
As the FAST detector is not scale-invariant, in order to ensure approximate scale invariance, feature points can be detected from an image pyramid~\cite{rub_iccv11}, which is called the multi-scale FAST detector.

\textbf{The AGAST descriptor}~\cite{mai_eccv10}, an acronym for Adaptive and Generic Accelerated Segment Test, is an extension of the FAST detector.
There are two major improvements in the AGAST descriptor.
The first one is the extension of the configuration space.
In the AGAST descriptor, two additional types of the surrounding pixels are added in order to the configuration space: \textit{not brighter} and \textit{not darker}.
By doing so, a more efficient decision tree can be constructed.
The second improvement is that the AGAST descriptor adaptively switches two different decision trees according to the probability of a pixel state to be similar
to the nucleus.

In \cite{leu_iccv11}, the multi-scale version of the AGAST detector is used.
Local features are first detected from multiple scales, and then non-maxima suppresion is performed in scale-space according to the FAST score.
Finally, scales and positoins of the detected local features are refined in a similar way to the SIFT detector.

\subsection{Feature Descriptors}
\label{sec:desc}

\subsubsection{Differential Invariants Descriptor}
Differential invariants descriptor was used in the pioneer work of local feature-based image retrieval~\cite{sch_tpami97}.
It consists of components of local jets~\cite{koe_bc87} and has rotation invariance:
\begin{equation}
v =
\begin{bmatrix}
L \\
L_i L_i \\
L_i L_{ij} L_j \\
L_{ii} \\
L_{ij} L_{ji} \\
\epsilon_{ij} (L_{jkl} L_i L_k L_l - L_{jkk} L_i L_l L_l \\
L_{iij} L_j L_k L_k - L_{ijk} L_i L_j L_k \\
-\epsilon_{ij} L_{jkl} L_i L_k L_l \\
L_{ijk} L_i L_j L_k \\
\end{bmatrix},
\end{equation}
where $i, j, k, l \in \{x, y \}$, and $L_x$ represents the convolution of image $I$ with the Gaussian derivative $G_x$ in terms of $x$ direction.

One approach to attain rotation-invariant local features is to adopt a scale-invariant descriptor like this differential invariants descriptor.
However, this approach results in less distinctive feature vector because it discards image information so that the resulting vector becomes the same irrespective of the degree of rotation.
Therefore, many descriptors adopts an orientation estimation step, and then feature descriptors extracts (scale-variant) feature vecctors relative to this orientation and therefore achieve invariance.

\subsubsection{SIFT Descriptor}
The SIFT~\cite{low_cvpr99, low04} descriptor is one of the most widely used feature descriptors, and sometimes combined with the other detectors (e.g. the Harris/Hessian-Affine detectors) as well as the SIFT detector.
In the SIFT descriptor, the orientation of local region $(x, y, \sigma)$ is estimated before description as follows.
Firstly, the gradient magnitude $m(x, y)$ and orientation $\theta(x, y)$ are computed using pixel differences:
\begin{align}
m(x, y) =& \biggl( \bigl( L(x + 1, y) - L(x - 1, y) \bigr)^2 \\
& + \bigl( L(x, y + 1) - L(x, y - 1) \bigr)^2 \biggr)^{1/2}, \\
\theta(x, y) =& \tan^{-1} \frac{L(x, y + 1) - L(x, y - 1)}{L(x + 1, y) - L(x - 1, y)},
\end{align}
where $L(x, y)$ denotes the intensity at $(x, y)$ in the image $I$ smoothed by the Gaussian with the scale parameter corresponding to the detected region.
Then, an orientation histogram is formed from the gradient orientations of sample pixels within the feature region;
the orientation histogram has 36 bins covering the 360 degree range of orientations.
Each pixel votes a score of the gradient magnitude $m(x, y)$ weighted by a Gaussian window to the bin corresponding to orientation $\theta(x, y)$.
The highest peak in the histogram is detected, which corresponds to the dominant direction of local gradients.
If any, the other local peaks that are within 80\% of the highest peak are used to create local features with that orientations~\cite{low04}.

After the assignment of the orientation, the SIFT descriptors are computed for normalized image patches.
The descriptor is represented by a 3D histogram of gradient location and orientation, where location is quantized into a $4 \times 4$ location grid and
the orientation is quantized into eight bins, resulting in the 128-dimensional descriptor.
For each of sample pixels, the gradient magnitude $m(x, y)$ weighted by a Gaussian window is voted to the bin corresponding to $(x, y)$ and $\theta(x, y)$ similar to the orientation estimation.
In order to handle a small shift, a soft voting is adopted, where scores weighted by trilinear interpolation are additionally voted to seven neighbor bins (voted to eight bins in total).
Finally, the feature vector is $\ell_2$ normalized to reduce the effects of illumination changes.

It is shown that certain post processing improves the discriminative power of the SIFT descriptor~\cite{ara_cvpr12, kob_cvpr14}.
In \cite{ara_cvpr12}, it is proposed to transform the SIFT descriptors by (1) $\ell_1$-normalization of the SIFT descriptor instead of $\ell_2$ and (2) taking square root each dimension.
The resulting descriptor is called RootSIFT.
Comparing RootSIFT using $ell_2$ distance correspond to using the Hellinger kernel in comparing the original SIFT descriptors.
In \cite{kob_cvpr14}, explict feature map of the Dirichlet Fisher kernel is proposed to transform the histogram-based feature vector (including the SIFT descriptor) to more discriminative one.

\subsubsection{SURF Descriptor}
The orientation assignment of the SURF Descriptor~\cite{bay_eccv06, bay_cviu08} is similar to the SIFT descriptor.
While the gradient magnitude and orientation are calculated from the image smoothed by the Gaussian in the SIFT descriptor, the Haar-wavelet responses in $x$ and $y$ directions are used in the SURF descriptor, where integral images are used for efficient calculation of the Haar-wavelet response.
Letting $s$ denote the characteristic scale of the SURF feature, the size of the Haar-wavelet is set to $4s$.
The Haar-wavelet responses of the pixels in a circular with the radius of $6s$ are accumulated using a sliding window with the size of $\pi/3$ and the dominant orientation is obtained.

In description, the feature region is first rotated using the estimated orientation, and divided into $4 \times 4$ subregions.
For each of the subregions, $d_x$, $d_y$, $|d_x|$, and $|d_y|$ are computed at $5 \times 5$ regularly spaced sample points, where $d_x$ and $d_y$ are the Haar-wavelet responses with the size of $2s$ in $x$ and $y$ directions.
These values are accumulated with the Gaussian weights, resulting in a subvector $v = (\sum d_x, \sum d_y, \sum |d_x|, \sum |d_y|)$.
The subvectors of $4 \times 4$ regions are concatenated to form the 64-dimensional SURF descriptor.


\subsubsection{BRIEF Descriptor}
The Binary Robust Independent Elementary Features (BRIEF) descriptor~\cite{cal_eccv10} is a pioneering work in the area of recent binary descriptors~\cite{hei_eccv12}.
Binary descriptors are quite different from the descriptors discussed above because they extract binary strings from patches of interest regions for efficiency instead of extracting gradient-based high-dimensional feature vectors like SIFT.
The distance calculations between binary features can be done efficiently by XOR and POPCNT operations.

The BRIEF descriptor is a bit string description of an image patch constructed from a set of binary intensity tests.
Many binary descriptors utilize similar binary tests in extracting binary strings.
Consider the $t$-th smoothed image patch $p_t$,
a binary test $\tau$ for $d$-th bit is defined by:
\begin{equation}
\label{eq:binary_test}
	x_{td} = \tau(p_t; a_d, b_d) =
	\begin{cases}
		\, 1 & \mathrm{if} \; p_t(a_d) \ge p_t(b_d) \\
		\, 0 & \mathrm{else}
	\end{cases},
\end{equation}
where
$a_d$ and $b_d$ denote relative positions in the patch $p_t$,
and $p_t(\cdot)$ denotes the intensity at the point.
Using $D$ independent tests,
we obtain $D$-bit binary string $x_t = (x_{t1}, \cdots, x_{td}, \cdots, x_{tD})$ for the patch $p_t$.
In the original BRIEF descriptor~\cite{cal_eccv10}, the relative positions $\{ (a_d, b_d) \}_d$ are randomly selected from certain probabilistic distributions (e.g. Gaussian).

\textbf{The ORB descriptor} \cite{rub_iccv11} is a modified version of the BRIEF descriptor, where two improvements are proposed: a orientation assignment and a learning method to optimize the positions $\{ (a_d, b_d) \}_d$.
In the orientation assignment, the intensity centroid \cite{ros_cviu99} is used.
The intensity centroid $C$ is defined as
\begin{equation}
C = \left( \frac{m_{10}}{m_{00}}, \frac{m_{01}}{m_{00}} \right),
\end{equation}
where $m_{pq}$ is the moments of the feature region:
\begin{equation}
m_{pq} = \sum_{x,y} x^p y^q I(x, y).
\end{equation}
The orientation $\theta$ of the vector from the center of the feature region to $C$ is obtained as:
\begin{equation}
\theta = \mathrm{atan2}(m_{01}, m_{10}).
\end{equation}

In athe learning method, the positions $\{ (a_d, b_d) \}_d$ are optimized so that the average value of each resulting bit is close to 0.5, and bits are not correlated.
This is achieved by an algorithm which greedy chooses the best binary test from all possive tests:
\begin{enumerate}
\item Calculate means of bits of all binary tests using traning patches rotated by $\theta$.
\item Sort the bits according to their distance from a mean of 0.5. Let $T$ denote the resulting vector.
\item Initialize the result vector $R$ by the first test in $T$ and remove it from $T$.
\item Take the next test from T, and compare it against all tests in $R$.
If its absolute correlation is smaller than a threshold, discard it; else add it to R.
Repeat this step until there are 256 tests in $R$.
If there are fewer than 256, raise the threshold and try again.
\end{enumerate}
This descriptor is usually combined with the multi-scale FAST detector, and therefore coined as Oriented FAST and Rotated BRIEF (ORB).

\textbf{The BRISK descriptor}~\cite{leu_iccv11}, an acronym for Binary Robust Invariant Scalable Keypoints, is a binary fescriptor similar to the ORB descriptor but proposed at the same time.
The major difference from the ORB descriptor is that the BRISK descriptor utilizes different sampling patterns for binary tests.
The BRISK sampling pattern is defined by the locations $p_i = (a_i, b_i)$ equally spaced on circles concentric with the keypoint, similar to the DAISY\cite{win_cvpr07, tol_pami10} descriptor.
For each location, the responses of different sizes of Gaussian kernel $I(p_j, \sigma)$ and $I(p_i, \sigma_i)$ at two different points $p_i$ and $p_j$ are compared in the tests as done in Eq.~(\ref{eq:binary_test}), while the intensities at two different points of smoothed image are compared in the ORB descriptor.
In the BRISK descriptor, sampling-point pairs whose distances are shorter than a threshold are used for description, resulting in the 512-bit descriptor.
The oriention $\tan^{-1}(g)$ is also estimated using the sampling pattern:
\begin{align}
g(p_i, p_j) = (p_i - p_j) \frac{I(p_j, \sigma_j) - I(p_i, \sigma_i)}{|| p_j - p_i ||^2}, \\
g = \begin{bmatrix}
g_x\\
g_y\\
\end{bmatrix}
=\frac{1}{L} \sum_{(p_i, p_j) \in \mathcal{L}} g(p_i, p_j),
\end{align}
where $\mathcal{L}$ is sampling-point pairs whose distances are relatively long.

\textbf{The FREAK descriptor}~\cite{ala_cvpr12}, an acronym for Fast REtinA Keypoint, is a binary fescriptor similar to the ORB and FREAK descriptor.
The FREAK sampling pattern mimics the \textit{retinal ganglion cells} distribution with their corresponding receptive fields, resulting in the very similar sampling pattern fo the DAISY\cite{win_cvpr07, tol_pami10} descriptor.
In the FREAK descriptor, the responses of different sizes of Gaussian kernel at two different points are compared in the tests similar to the BRISK descriptor, but the learning method in the ORB descriptor is used to select the effective binary tests.
The orientation is also calculated in a similar way to the BRISK descriptor.
The differences are the sampling pattern and the number of point pairs.
The number of point pairs is reduced to 45, achieving smaller memory requirement.

\section{Image Representations}
\label{sec:image_representation}

\subsection{Bag-of-Visual Words}
The BoVW framework is the de-facto standard way to encode local features into a fixed length vector.
The BoVW framework is firstly proposed in the context of object matching in videos~\cite{siv03}, it has been used in various tasks in image retrieval~\cite{nis06, phi07, jeg08}, image classification~\cite{csu_slcv04, laz_cvpr06, jia_civr07}, and video copy detection tasks~\cite{dou10b, uch_icmr11}.
In the BoVW framework, represantative vectors called visual words (VWs) or visual vocabulary are created.
These representative vectors are usually created by applying $k$-means algorithm to training vectors, and resulting centroids are used as VWs.
Feature vectors extracted from an image are quantized into VWs, resulting in a histogram representation of VWs.
Image (dis)similarity is measured by $\ell_1$ or $\ell_2$ distance between the normalized histograms.

\begin{figure*}[tb]
	\centering
	\includegraphics[width=\linewidth]{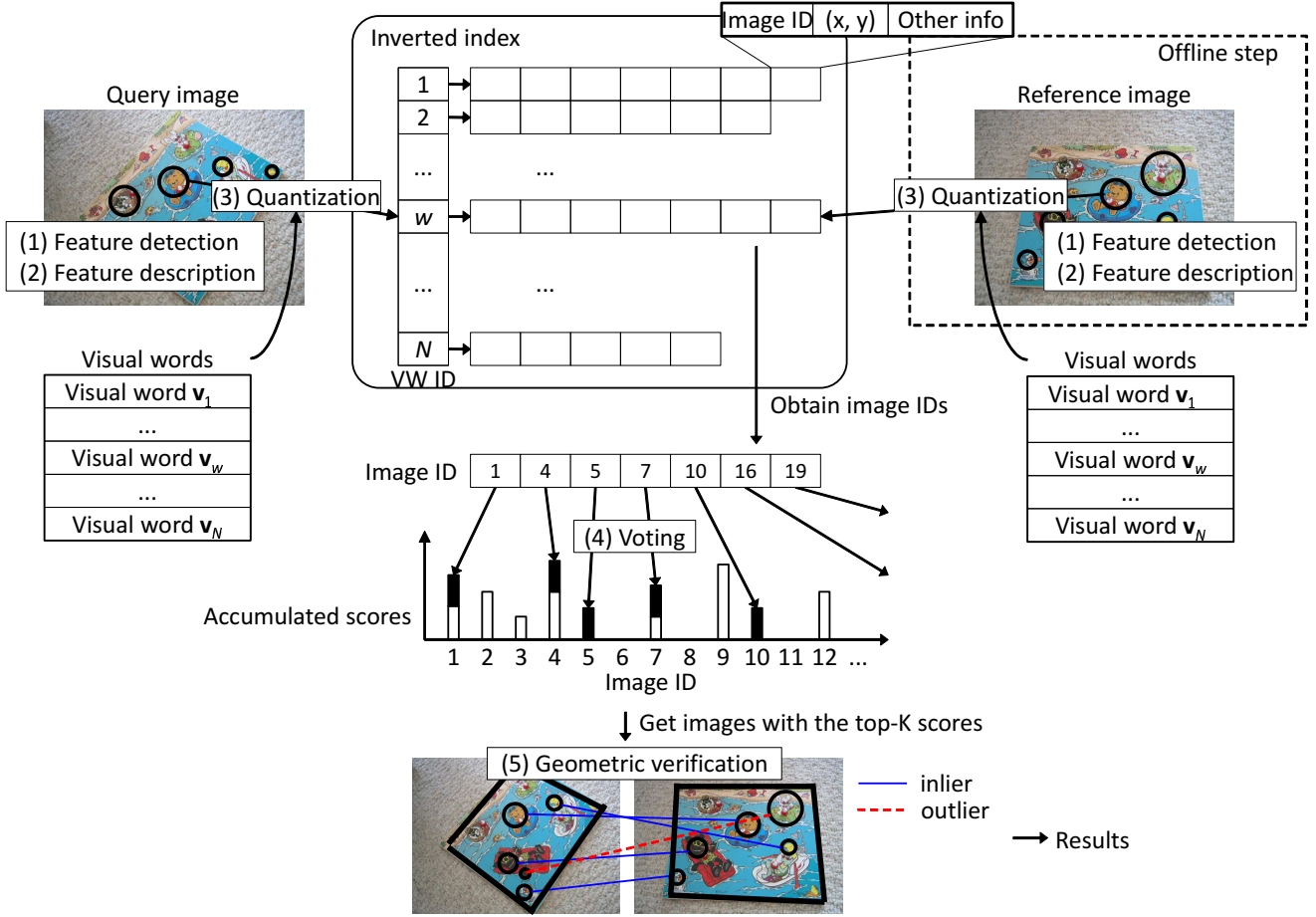}
	\caption{A framework of image retrieval using the inverted index data structure.}
	\label{fig:framework}
\end{figure*}

As the histograms are generally sparse\footnote{Note that, in image classification tasks, the BoVW histogram is often not sparse but dense because extremely larger number of features are extracted with dense grid sampling and the number of VWs is relatively small.
Therefore, it is not standard to use inverted index but simply treat BoVW histogram as a dense vector.}, an inverted index and a voting function enables an efficient similarity search~\cite{siv03}.
Figure~\ref{fig:framework} shows a framework of image retrieval using the inverted index data structure.
The inverted index contains a list of containers for each VW, which store information of reference features such as the identifiers of reference images, the positions $(x, y)$ of the reference features, or other information used in search step.

The framework involves three steps: training, indexing, and search steps.
In training step, VWs are trained by performing the $k$-means algorithm to training vectors. Other trainings requried for indexing or search are done, if any.
In indexing step, feature regions are firstly detected in a reference image, and then feature vectors are extracted to describe these regions.
Finally, each of these reference feature vectors is quantized into VW, and the identifier of the reference image is stored in the corresponding lists with other metadata related the reference feature.
In search step, feature regions are detected in a query image, feature vectors are extracted, and these query feature vectors are quantized into VWs in the same manner as done in the indexing step.
Then, each of query feature vote a certain score to reference images whose identifiers are found in the corresponding lists.
A Term Frequency-Inverse Document Frequency (TF-IDF) scoring~\cite{siv03} is often used in voting function.
The voting scores are accumulated over all of the query feature features, resulting similarities between the query image and the reference images.
The results obtained in voting function optionally refined by Geometric Verification (GV) or spatial re-ranking~\cite{phi07, chu_iccv07}, which will be described later.

The BoVW is the most widely used framework in local feature-based image retrieval, and therefore many extensions of the BoVW framework are proposed.
In the following, we comprehensively review these extensions.
We classify the BoVW extensions into the following groups in this paper: large vocabulary, multiple assignment, post-filtering, weighted voting, geometric verification, weak geometric consistency, and query expansion.
These are reviewed one by one.


\subsubsection{Large Vocabulary}
Using a large vocabulary in quantization, e.g. one million VWs, increases discriminative power of VWs, and thus improves search precision.
In \cite{nis06}, it is proposed to quantize feature vectors using a vocabulary tree, which is created by hierarchical $k$-means clustering (HKM) instead of a flat $k$-means clustering.
The vocabulary tree enables extremely efficient indexing and retrieval while increasing discriminative power of VWs.
A hierarchical TF-IDF scoring is also proposed to alleviate quantization error caused in using a large vocabulary.
This hierarchical scoring can be considered as a kind of multiple assignment explained later.

In \cite{phi07}, approximate $k$-means (AKM) is proposed to create large vocabulary.
AKM is an approximated version of $k$-means algorithm, where an approximate nearest neighbor search method is used in assigning training vectors to their nearest centroids.
In AKM, a forest of randomized $k$-d trees~\cite{ami_nc97, lep_cvpr05, sil08} is used for approximate nearest neighbor search, where are the randomized $k$-d trees are simultaneously searched using a single priority queue in a best-bin-first manner~\cite{bei_cvpr97}.
This nearest neighbor search is performed in quantization as well as in clustering.
It is shown that AKM outperforms HKM in terms of image search precision.
This is because HKM minimizes quantization error only locally at each node while the flat $k$-means minimizes total quantization error, and AKM successfully approximate the flat $k$-means clustering.

In \cite{mik_eccv10, mik_ijcv13}, the combination of the above HKM and AKM, namely approximate hierarchical $k$-means (AHKM), is proposed to construct further larger vocabulary.
The AHKM tree consists of two levels, where each level has 4K nodes.
The first level is constructed by AKM using randomly sampled training vectors.
Then, over 10 billion training vectors are devided into 4K clusters by using the first level centroids.
For each of the above 4K clusters AKM is further applied to construct the second level with 4K centroids, resulting in 16M VWs.
In the construction of the first level, th tree structure is balanced so that averaging the speed of the retrieval~\cite{jeg_ijcv10, tav_cbmi11}.

\subsubsection{Multiple Assignment}
\begin{figure}[tb]
	\centering
	\includegraphics[width=0.99\linewidth]{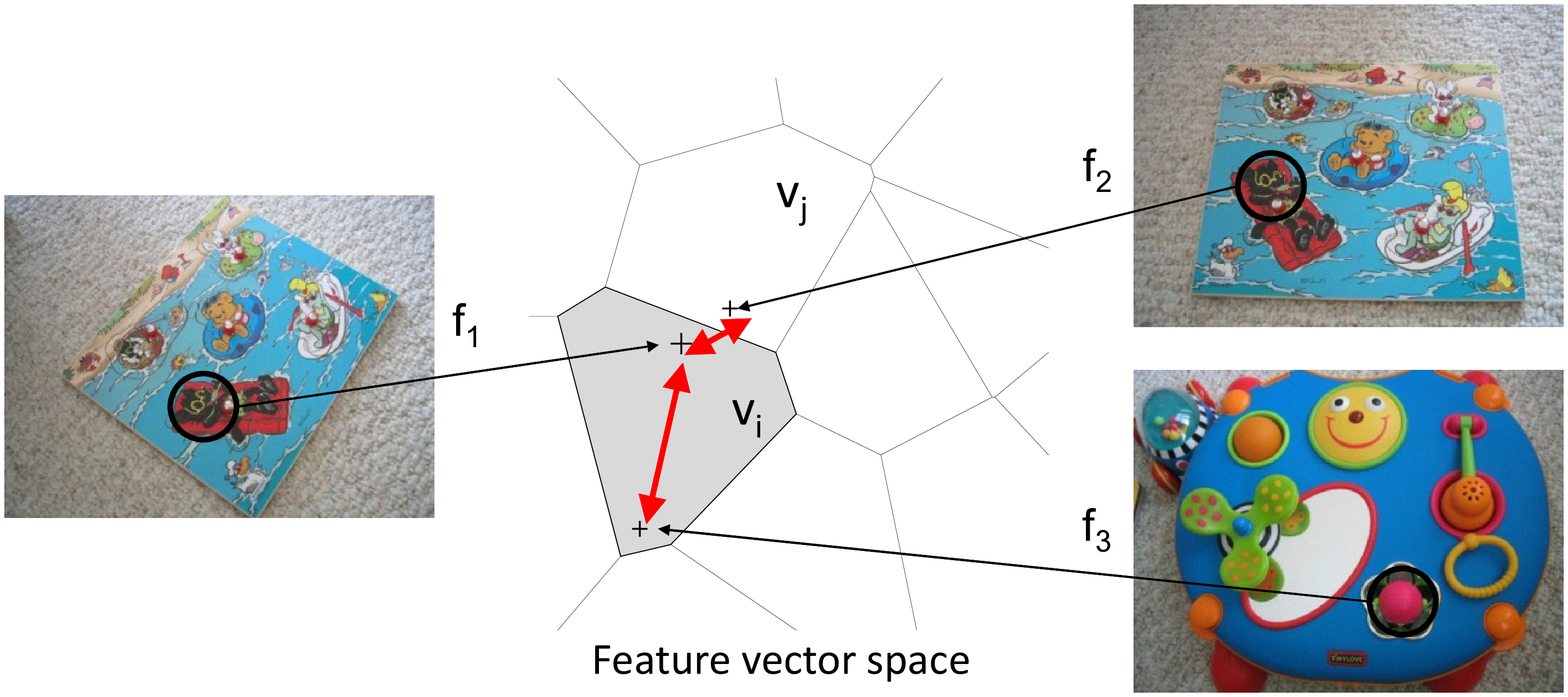} \\
	(a) Problems in the BoVW framework. \\
	\begin{minipage}[b]{0.49\linewidth}
		\centering
		\includegraphics[width=\linewidth]{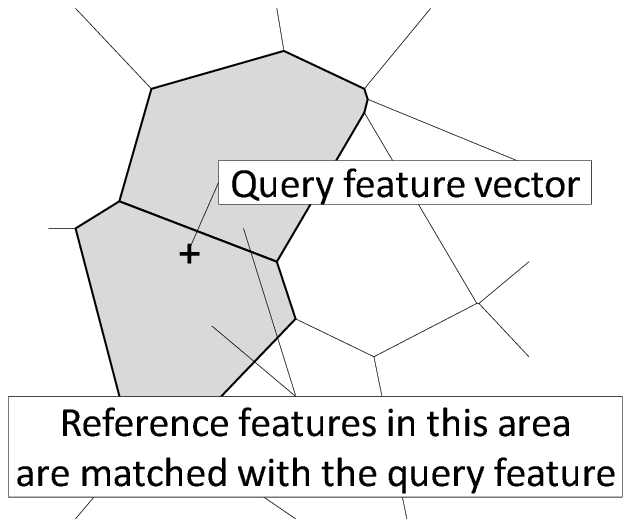} \\
		(b) Improving matching accuracy by multiple assignment. \\
	\end{minipage}
	\begin{minipage}[b]{0.49\linewidth}
		\centering
		\includegraphics[width=\linewidth]{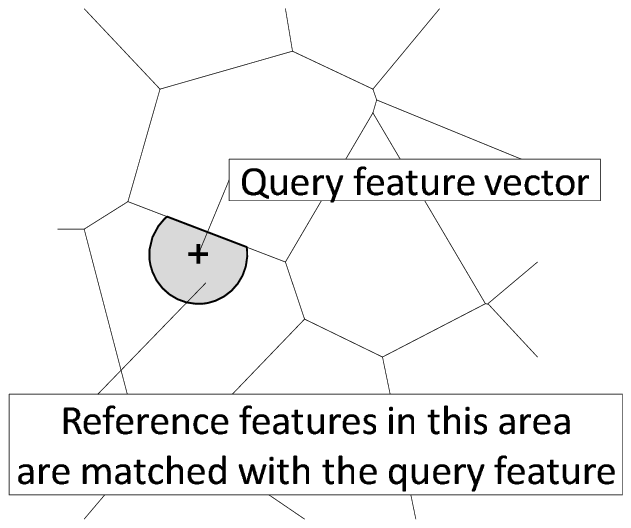} \\
		(c) Improving matching accuracy by filtering approach. \\
	\end{minipage}
	\caption{Problems in the BoVW framework and its solutions.}
	\label{fig:qerror}
\end{figure}

One significant drawback of VW-based matching is that two features are matched if and only if they are assigned to the same VW.
Figure~\ref{fig:qerror} illustrates this drawback.
In Figure~\ref{fig:qerror} (a), two features $f_1$ and $f_2$ extracted from the same object are close to each other in the feature vector space.
However, there are the boundary of the Voronoi cells defined by VWs, and they are assigned to the different VWs $v_1$ and $v_j$.
Therfore,  $f_1$ and $f_2$ are not matched in the naive BoVW framework.
Multiple assignment (or soft assignmet) is proposed to solve this problem.
The basic idea is to assign feature vectors not only to the nearest VW but to the several nearest VWs.
Figure~\ref{fig:qerror} (b) explains how it works.
Suppose $f_1$ is a query vector and assigned to the nearest two VWs, $v_i$ and $v_j$.
In this case, reference features inclusing $f_2$ in the gray area are matched to $f_1$.
In general, multiple assignment improves recall of matching features while degrading precision because each feature is matched with larger number of features in the database compared with hard (single) assignment case.

In \cite{phi_cvpr08}, each of reference features is assigned to the fixed number $r$ of the nearest VWs in indexing and the corresponding score $\exp -\frac{d^2}{2\alpha^2}$ is additionally stored in the inverted index, where $d$ is the distance from the VW to the reference feature, and $\alpha$ is a scaling parameter.
It is shown that the multiple assignment brings a considerable performance boost over \textit{hard-assignment}~\cite{phi07}.
This multiple assignment is called \textit{reference-side} multiple assignment because it is done in indexing.
Beucase reference-side multiple assignment increases the size of the index almost proportionally to the factor $r$, the following \textit{query-side} multiple assignment is often used.
In \cite{phi_cvpr08}, it is also proposed to perform multiple assignment against image patch, namely \textit{image-space} multiple-assignment.
In \textit{image-space} multiple-assignment, a set of descriptors is extracted from each image patch by synthesizing deformations of the patch in the image space and assign each descriptor to the nearest visual word.
However, it is shown that, compared to descriptor-space soft-assignment explained above, image-space multiple assignment is much more computationally expensive while not so effective.

In \cite{jeg_ijcv10}, it is proposed to perform multiple assignment to a query feature (query-side multiple assignment), where the distance $d_0$ to the nearest VW from a query feature is used to determine the number of multiple assignments.
The query feature is assigned to the nearest VWs such that the distance to the VW is smaller than $\alpha d_0$ ($\alpha = 1.2$ in \cite{jeg_ijcv10}).
This approach adaptively changes the number of assigned VWs according to ambiguity of the feature.

While all of the above methods utilize the Euclidean distance in selecting VWs to be assigned, in \cite{mik_eccv10, mik_ijcv13}, it is proposed to exploit a probabilistic relationships $P(W_j | W_q)$ of VWs in multiple assignment.
$P(W_j | W_q)$ is the probability of observing VW $W_j$ in a reference image when VW $W_q$ was observed in the query image.
In other words, $P(W_j | W_q)$ represents which other VWs (called alternative VWs) that are likely to contain descriptors of matching features.
The probability is learnt from a large number of matching image patches.
For each VW $W_q$, a fixed number of alternative VWs that have the
highest conditional probability $P(W_j | W_q)$ is recorded in a list and used in multiple assignment;
a query feature assigned to the VW $W_q$, it is also assinged to the VWs in the list.

\subsubsection{Post-filtering}
As the naive BoVW framework suffers from many false matches of local features, post-filtering approaches are proposed to eliminate unreliable feature matches.
In post-filtering approaches, after VW-based matching, matched feature pairs are further filtered out according to the distances between them.
For example, in Figure~\ref{fig:qerror}, two features $f_1$ and $f_3$ are far from each other in feature vector space but in same the Voronoi cell, thus they are matched in the naive BoVW framework.
In Figure~\ref{fig:qerror} (c), post-filtering is applied;
the query feature is matched with only the reference features in the gray area, flitering out the feature $f_3$.
Post-filtering approches have similar effect as using a large vocabulary because both of them improve accuracy of feature matching.
While post-filtering approaches try to improve the precision of feature matches with only slight degradation of recall, simply using a large vocabulary causes a considerable degradation of recall in feature matching~\cite{jeg08}.

In post-filtering approaches, after VW-based matching, distances between a query feature and reference features that are assigned to the same VW should be calculated for post-filtering.
However, as exact distance calculation is undesirable in terms of computational cost and memory requirement to store raw feature vectors.
Therefore, in \cite{jeg08, jeg_ijcv10, zha_tmm10, wan_tomm14}, feature vectors extracted from reference images are encoded into binary codes (typically 32-128 bit codes)
via random orthogonal projection followed by thresholding for binarizing projected vectors.
While all VWs share a single random orthogonal matrix,
each VW has individual thresholds so that feature vectors are binarized into 0 or 1 with the same probability.
These codes are stored in an inverted index with image identifiers (sometimes with other information on the features.
In a search step, after VW-based matching, Hamming distances between codes of query and matched reference features are calculated.
Matched features with larger Hamming than a predefined threshold are filtered out,
which considerably improves the precision of matching with only slight degradation of recall.

In \cite{jeg10, uch_icme12, qin_cvpr13}, a product quantization-based method is proposed and
shown to outperform other short codes like spectral hashing (SH)~\cite{wei08} or a transform coding-based method~\cite{bra_cvpr10}
in terms of the trade-off between code length and accuracy in approximate nearest neighbor search.
In the PQ method, a reference feature vector is decomposed into low-dimensional subvectors.
Subsequently, these subvectors are quantized separately into a short code, which is composed of corresponding centroid indices.
The distance between a query vector and a reference vector is approximated by the distance between a query vector and the short code of a reference vector.
Distance calculation is efficiently performed with a lookup table.
Note that the PQ method directly approximates the Euclidean distance between a query and reference vector, while the Hamming distance obtained by the HE method only reflects their similarity.
In \cite{uch_gcce14}, the filtering approach is extended to recent binary features, where informative bits are selected for each VW and are stored in the inverted index in order to perform the post-filtering.

\subsubsection{Weighted Voting}
In voting function, TF-IDF scoring~\cite{siv03} is often used.
Some researches try to improve the image retrieval accuracy by modifying this scoring.
One directon to do this is the modification of the standard IDF.
In \cite{zhe_cvpr13, zhe_tip14b}, $\ell_p$-norm IDF is proposed, which can be considered as a generalized version of the standard IDF.
The standard IDF weight for the visual word $z_k$ is defined as:
\begin{equation}
IDF(z_k) = \log \frac{N}{n_k},
\end{equation}
where $N$ denotes the number of images in the database and $n_k$ denotes the number of images that contain $z_k$.
The $\ell_p$-norm IDF is defined as:
\begin{equation}
pIDF(z_k) = \log \frac{N}{\sum_{I_i \in P_k} w_{i,k} v_{i,k}^p},
\end{equation}
where $v_{i,k}$ denotes the occurrences of $z_k$ in the image $I_i$ and $w_{i,k}$ is normalization term.
It is reported that when $p$ is about 3-4, $\ell_p$-norm IDF achieves better accuracy than the standard IDF.
In \cite{mur_tmm14}, BM25 with exponential IDF weights (EBM25) is proposed.
BM25 is a ranking function used for document retrieval and it includes the IDF term in its definition.
This IDF term is extended to the exponential IDF that is capable of suppressing the effect of background features.
In \cite{uch_cva13}, theoretical scoring method is derived by formulating the image retrieval problem as a maximum-a-posteriori estimation.
The derived score can be used an alternative to the standard IDF.

The distances between the query and reference features obtained in the post-filtering approach are aften exploited in weighted voting.
In \cite{jeg_cvpr09}, the weight is calculated as a Gaussian function of a Hamming distance between the query and reference vector.
In \cite{jeg_ijcv10}, the weight is calculated based on the Hamming distance between the query and reference vector and the probability mass function of the binomial distribution.
In \cite{jai_icmr12}, the weight is calculated based on rank information because a rank criterion is used in post-filtering in the literature, while in \cite{uch_icme12}, the weight is calculated based on ratio information.

\subsubsection{Geometric Verification}
Geometric Verification (GV) or spatial re-ranking is important step to improve the results obtained by voting function~\cite{phi07, chu_iccv07}.
In GV, transformations between the query image and the top-$R$ reference images in the list of voting results are estimated, eliminating matching pairs which are not consistent with the estimated transformation.
In the estimation, the RANdom SAmple Consensus (RANSAC) algorithm or its variants~\cite{chum_accv04, chu_cvpr05} are used.
Then, the score is updated counting only inlier pairs.
As a transformation model, affine or homography matrix is usually used.

In \cite{low04} a 4 Degrees of Freedom (DoF) affine transformation is estimated in two stages.
First, a Hough scheme estimates a transformation with 4 parameters; 2D location, scale, and orientation.
Each pair of matching regions generates these parameters that \textit{vote} to a 4D histogram.
In the second stage, the sets of matches from a bin with at least 3 entries are used to estimate a finer 2D affine transform.
In \cite{phi07}, three affine sub-groups for hypothesis generation are compared, with DoF ranging between 3 and 5.
It is shown that 5 DoF outperforms the others but the improvement is small.
Because affine-invariant Hessian regions~\cite{mik_ijcv04} are used in \cite{phi07}, each hypothesis of even 5 DoF affine transformation can be generated from only a single pair of corresponding features, which greatly reduces the computational cost of GV.

The above method utilizes local geometry represented by an affine covariant ellipse in GV.
Because storing the parameters of ellipse regions significantly increases memory requirement, a method is proposed to learn discretized local geometry representation by minimizing average reprojection error in the space of ellipses in \cite{per_cvpr09}.
It is shown that the representation requires only 24 bits per feature without drop in performance.

\subsubsection{Weak Geometric Consistency}
Geometric verification explained above is very effective but costly.
Therefore, it is only applicable up to a few hundred images.
To overcome this problem, Weak Geometric Consistency (WGC) method is proposed in \cite{jeg08, jeg_ijcv10}, where WGC filters matching features that are not consistent in terms of angle and scale.
This is done by estimating rotation and scaling parameters between a query image and a reference image separately assuming the following transformation:
\begin{equation}
\label{eq:wgc_model}
\begin{bmatrix}
x_q \\
y_q\\
\end{bmatrix}
= s \times
\begin{bmatrix}
\cos \theta	& - \sin \theta\\
\sin \theta	& \cos \theta\\
\end{bmatrix}
\times
\begin{bmatrix}
x_p \\
y_p\\
\end{bmatrix}
+
\begin{bmatrix}
t_x \\
t_y\\
\end{bmatrix},
\end{equation}
where $(x_q, y_q)^{\top}$ and $(x_p, y_p)^{\top}$ are the positions in query and reference image, $s$ and $\theta$ are scaling and rotation parameters, and $(t_x, t_y)^{\top}$ is translation.
In order to efficiently estimate the scaling and rotation parameters, each of feature matches votes to 1D histograms of angle differences and log-scale differences.
The score of the largest bin among these two 1D histograms is used as image similarity.
This scoring reduces the scores of the images for which the points are not transformed by consistent angles and scales, while a set of points consistently transformed will accumulate its votes in the same histogram bin, keeping a high score.

In WGC, the log-scale difference histogram always has a peak corresponding to 0 (same scale) because most of feature detectors utilizes image pyramid, resulting that most features are detected with the smallest scale.
To solve this problem, in \cite{zha_tmm10}, the absolute value of the translation $(t_x, t_y)^{\top}$ is estimated by voting instead of scaling and rotation parameters.
In \cite{zha_cvpr11, she_cvpr12, wan_mm13, zho_tmm15}, similarly, the translation $(t_x, t_y)^{\top}$ is estimated using 2D histogram instead of shrinking to 1D histogram of its absolute value.
Because the scaling and orientaton parameters in Eq.~(\ref{eq:wgc_model}) are not considered in \cite{zha_cvpr11}, the method proposed is not scale and rotation invariant.

In \cite{ngo_mm06}, a new measure called Pattern Entropy (PE) is introduced, which measures the coherency of symmetric feature matching across the space of two images similar to WGC.
This coherency is captured with two histograms of matching orientations, which are composed of the angles formed by the matching lines and horizontal or vertical axis in the synthesized image where two images are aligned horizontally and vertically.
As PE is not scale nor rotation invariant, the improved version of PE, namely Scale and Rotation invariance PE (SR-PE), is proposed in \cite{zha_tip09}.
In SR-PE, rotation and scaling parameters are estimated similar to WGC.
However, in SR-PE, these parameters are estimated using two pairs of matching features because it does not utilize scale and orientation parameters of local features.
Therefore, it is not applicable to all reference images.

In \cite{tsa10}, three types of scoring methods based on weak geometric information are proposed for re-ranking: location geometric similarity scoring, orientation geometric similarity scoring, scale geometric similarity scoring.
The orientation and scale geometric similarity scorings are the same as WGC~\cite{jeg08, jeg_ijcv10}.
The location geometric similarity scoring is calculated by transforming the location information into distance ratios to measure the geometric similarity;
each of all prossible combination of two matching pairs votes a score to 1D histogram, where each bin is defined by the log ratio of the distances of two points in a query image and corresponding two points in a reference image.
The geometric similarity is defined by the score of the bin with miximum votes.
The location geometric similarity scoring cannot be integrated with inverted index and is only applicable to re-ranking beucase it requires all combination of matching features.

In \cite{cao_cvpr10} spatial-bag-of-features representation is proposed, which is a generalization of the spatial pyramid \cite{laz_cvpr06, lin_eccv10}.
The spatial pyramid is not invariant to scale, rotation, nor translation.
Spatial-bag-of-features utilizes a specific calibration method which re-orders the bins of a BoVW histogram in order to deal with the above transformations.
In \cite{wu_cvpr09, wu_iccv09, zha_mm10}, contextual information is introduced to the BoVW framework by bundling multiple features~\cite{wu_cvpr09, zha_mm10} or extracting feature vectors in multiple scales~\cite{wu_iccv09}.

\subsubsection{Query Expansion}
In the text retrieval literature a standard method for improving performance is query expansion, where a number of the highly ranked documents are integrated into a new query.
By doing so, additional information can be added to the original query, resulting better search precision.
Because the idea of the BoVW framework is comes from the Bag-of-Words (BoW) in text retrieval, it is also natural to borrow query expansion from text retrieval.
In \cite{chu_iccv07}, various types of query expansions are introduced and compared in the visual domain.
There are some insightful observations found in In \cite{chu_iccv07}.
Fistly, simply using the top $K$ results for expansion degrades search precision;
false positives in the top $K$ results make the expanded queries less informative.
Secondly, averaging geometrically verified results for expansion significantly improves the results because GV excludes false positives from results to the original query.
Furthermore, recursively performing this average query expansion further improves the results.
Finally, resolution expansion achieves the best performance, which first clusters geometrically verified results into groups, and then issues multiple expanded queries independently created from these groups.

In \cite{chu_cvpr11}, three approaches are proposed in order to improve query expansion:
automatic tf-idf failure recovery, incremental spatial reranking, and context query expansion.
In automatic tf-idf failure recovery, after GV, if inlier ratio is smaller than threshold, noisy VWs called \textit{confuser} is estimated according to likelihood ratio.
Then, the original query is updated by removing confuser if it improves inlier ratio.
In incremental spatial reranking\footnote{Incremental spatial reranking is not a query expansion method, but a variant of GV (spatial reranking). Therefore, it is applicable even if query expansion is not used.}, instead of performimg GV to the top $K$ results using the original query, the original query is incrementally updated at each GV if sufficient number of inliers are found in the GV.
In context query expansion, a feature outside the bounding box~\footnote{Here, it is assumed a query consists of a query image and a bounding box representing the target object of the search.} is added to an expanded query if it is consistently found in multiple geometrically verified results.

In \cite{ara_cvpr12}, discriminative query expansion is proposed, where a linear SVM is trained using geometrically verified results as positive samples and results with lower scores in voting as negative samples.
The results are reranked according to the distances from the boundary of the trained SVM.

\subsubsection{Summary}
In this section, the BoVW extensions were grouped into seven types of approaches: large vocabulary, multiple assignment, post-filtering, weighted voting, geometric verification, weak geometric consistency, and query expansion.
These approaches are complementary to each other and often used together.
Table~\ref{tab:summary} summarizes the literatures in which these approaches are proposed.
We show which approaches are used in the literatures and summarize the best results on publicly available datasets: the UKB\footnote{http://vis.uky.edu/~stewe/ukbench/}, Oxford5k\footnote{http://www.robots.ox.ac.uk/~vgg/data/oxbuildings/}, Oxford105k, Paris\footnote{http://www.robots.ox.ac.uk/~vgg/data/parisbuildings/}, and the Holidays\footnote{http://lear.inrialpes.fr/~jegou/data.php} dataset.
Oxford105k consists of Oxford5k and 10k distractor images.
There are several observations through this summarization:
\begin{itemize}
\item Large vocabulary or post-filtering is adopted in all of the literatures. These approaches enhance the discreminative power of the BoVW framework and thus are essential for accurate image retrieval system.
\item Multiple assignment is also used in many literatures. This is because multiple assignment can improve the recall of feature-level matching at the cost of small increase of computational cost.
\item Geometric verification and query expansion are used in many literatures to boost the performance though geometric verification and query expansion are not the main proposal of these literatures. This is because geometric verification and query expansion are needed to achieve the state-of-the-art results on the publicly available datasets. Therefore, we think the absolute values of the accuracies are not directly reflecting the importances of the proposals.
\item In several literatures, visual words are learnt using the \textit{test} dataset. Visual words learnt on the test dataset tend to achieve significantly better accuracy than visual words learnt on an independent dataset. When considering the results, we should be aware of this. In Table~\ref{tab:summary}, we added '*' mark to the literatures in which visual words are learnt on the datasets.
\item While we did not specify the use of RootSIFT~\cite{ara_cvpr12}, RootSIFT has become a de-facto standard descriptor due to its effectiveness and simplicity.
\end{itemize}

\begin{table*}[tb]
	\centering
	\footnotesize
	\caption{Summary of existing literature. The use of seven types of approaches is specified: large vocabulary (LV), multiple assignment (MA), post-filtering (PF), weighted voting(WV), geometric verification (GV), weak geometric consistency (WGC), and query expansion (QE).
	For results, mean average precision (MAP) is shown for each literature and dataset as a performance measurement (higher is better).
	In addition to MAP, top-4 recall score is also shown for the UKB dataset (higher is better).}
	\label{tab:summary}
	\begin{tabular}{cc|ccccccc|cccccc} \hline
                  &      & \multicolumn{7}{c|}{Methods}                                                            & \multicolumn{6}{c}{Results}                      \\
Literature        &year  & LV         & MA         & PF         & WV         & GV         & WGC        & QE         & UKB   & UKB   & Ox5K  & Ox105K & Paris6K & Holiday \\ \hline
\cite{nis06}      & 2006 & \checkmark &            &            &            &            &            &            & 3.29  &       &       &        &         &         \\
\cite{phi07}      & 2007 & \checkmark & \checkmark &            &            & \checkmark &            &            &       &       & 0.645 &        &         &         \\
\cite{phi_cvpr08}*& 2008 & \checkmark & \checkmark &            &            & \checkmark &            & \checkmark &       &       & 0.825 & 0.719  &         &         \\
\cite{jeg_cvpr09} & 2009 &            & \checkmark & \checkmark & \checkmark & \checkmark & \checkmark &            & 3.64  & 0.930 & 0.685 &        &         & 0.848   \\
\cite{per_cvpr09}*& 2009 & \checkmark & \checkmark &            &            & \checkmark &            & \checkmark &       &       & 0.916 & 0.885  &         & 0.780   \\
\cite{jeg_ijcv10} & 2010 &            & \checkmark & \checkmark & \checkmark &            & \checkmark &            & 3.38  & 0.870 & 0.605 &        &         & 0.813   \\
\cite{mik_eccv10} & 2010 & \checkmark & \checkmark &            &            & \checkmark &            & \checkmark &       &       & 0.849 & 0.795  & 0.824   & 0.758   \\
\cite{zha_cvpr11} & 2011 & \checkmark &            &            &            & \checkmark & \checkmark &            &       &       & 0.713 &        &         &         \\
\cite{chu_cvpr11} & 2011 & \checkmark &            &            &            & \checkmark &            & \checkmark &       &       & 0.827 &        & 0.805   &         \\
\cite{qin_cvpr11} & 2011 & \checkmark &            &            &            &            &            & \checkmark &       &       & 0.814 & 0.767  & 0.803   &         \\
\cite{ara_cvpr12}*& 2012 & \checkmark & \checkmark &            &            & \checkmark &            & \checkmark &       &       & 0.929 & 0.891  & 0.910   &         \\
\cite{she_cvpr12}*& 2012 & \checkmark &            &            &            & \checkmark & \checkmark & \checkmark & 3.56  &       & 0.884 & 0.864  & 0.911   &         \\
\cite{zhe_cvpr13} & 2013 & \checkmark &            &            & \checkmark &            &            &            &       &       & 0.696 &        & 0.562   &         \\
\cite{qin_cvpr13} & 2013 &            & \checkmark & \checkmark & \checkmark &            &            & \checkmark &       &       & 0.850 & 0.816  & 0.855   & 0.801   \\
\cite{uch_cva13}  & 2013 &            & \checkmark & \checkmark & \checkmark &            &            &            & 3.55  & 0.910 &       &        &         &         \\
\cite{tol_iccv13} & 2013 &            & \checkmark & \checkmark & \checkmark & \checkmark &            & \checkmark &       &       & 0.804 & 0.750  & 0.770   & 0.810   \\
\cite{zhe_cvpr14} & 2014 &            & \checkmark & \checkmark &            &            &            &            & 3.71  &       &       &        &         & 0.840   \\
\cite{zho_tmm15}* & 2015 & \checkmark &            &            &            & \checkmark & \checkmark & \checkmark &       &       & 0.950 & 0.932  & 0.915   &         \\ \hline
	\end{tabular}
\end{table*}

\subsection{Fisher Kernel and Fisher Vector}

\subsubsection{Definition}
Fisher kernel is a powerful tool for combining the benefits of generative and discriminative approaches~\cite{jaa_nips98}.
Let $X$ denote a data item (e.g. a feture vector or a set of feature vector).
Here, the generation process of $X$ is modeled by a probability density function $p(X | \lambda)$ whose parameters are denoted by $\lambda$.
In~\cite{jaa_nips98},
it is proposed to describe $X$ by
the gradient $G^{X}_{\lambda}$ of the log-likelihood function,
which is also referred to as the Fisher score:
\begin{equation}
\label{eq:1}
G^{X}_{\lambda} =
\nabla_{\lambda} \mathcal{L} (X | \lambda),
\end{equation}
where $\mathcal{L} (X | \lambda)$ denotes the log-likelihood function:
\begin{equation}
\label{eq:likelihood}
\mathcal{L} (X | \lambda) = \log p(X | \lambda).
\end{equation}
The gradient vector describes the direction in which parameters should be modified to best fit the data~\cite{per_cvpr07}.
A natural kernel on these gradients is the Fisher kernel~\cite{jaa_nips98},
which is based on the idea of natural gradient~\cite{ama_nc98}:
\begin{equation}
K(X, Y) = G^{X}_{\lambda} F_{\lambda}^{-1} G^{Y}_{\lambda}.
\end{equation}
$F_{\lambda}$ is the Fisher information matrix of $p(X | \lambda)$ defined as
\begin{equation}
\label{eq:fim}
F_{\lambda} =
\mathrm{E}_X [ \nabla_{\lambda} \mathcal{L} (X | \lambda) \; \nabla_{\lambda} \mathcal{L} (X | \lambda)^{\mathrm{T}}].
\end{equation}
Because $F_{\lambda}^{-1}$ is positive semidefinite and symmetric,
it has a Cholesky decomposition $F_{\lambda}^{-1} = L_{\lambda}^{\mathrm{T}} L_{\lambda}$.
Therefore the Fisher kernel is rewritten as a dot-product between normalized gradient vectors $\mathcal{G}_{\lambda}^X$ with:
\begin{equation}
\label{eq:fv}
\mathcal{G}_{\lambda}^X = L_{\lambda} G^{X}_{\lambda}.
\end{equation}
The normalized gradient vector $\mathcal{G}_{\lambda}^X$ is referred to as the Fisher vector of $X$~\cite{per_eccv10}.

\subsubsection{GMM Fisher Vector}
In~\cite{per_cvpr07}, the generation process of feature vectors (SIFT) are modeled by the GMM, and the diagonal closed-form approximation of the Fisher vector is derived.
Then, the performance of the Fisher vector is significantly improved in \cite{per_eccv10} by using power-normalization and $\ell_2$ normalization.
The Fisher vector framework has achieved promising results and is becoming the new standard in both image classification~\cite{per_eccv10, san_ijcv13} and image retrieval tasks~\cite{per_cvpr10, jeg_cvpr10, jeg_pami12}.

Let $X = \{ x_1, \cdots, x_t, \cdots, x_T \}$ denote the set of low-level feature vectors extracted from an image.
and $\lambda = \{ w_i, \mu_i, \Sigma_i, i = 1..N \}$ denote the set of parameters for GMM with $N$ components.
From Eq.~(\ref{eq:likelihood}) and an independence assumption where $x_1, \cdots, x_T$ are independently generated,
wehave:
\begin{equation}
\mathcal{L} (X | \lambda) = \sum_{t=1}^{T} \log p(x_t | \lambda).
\end{equation}
The probability that $x_t$ is generated by GMM is:
\begin{equation}
p(x_t | \lambda) = \sum_{i =1}^N w_i p_i (x_t | \lambda).
\end{equation}
The $i$-th component $p_i$ is given by
\begin{equation}
p_i (x_t | \lambda) = \frac{\exp \left( -\frac{1}{2} (x - \mu_i)' \Sigma_i^{-1} (x - \mu_i) \right)}{(2\pi)^{D/2} |\Sigma_i|^{1/2}},
\end{equation}
where $D$ is the dimensionality of the feature vector $x_t$ and $| \cdot |$ denotes the determinant operator.
In~\cite{per_cvpr07}, it is assumed that the covariance matrices are diagonal because any distribution can be approximated with an arbitrary precision by a weighted sum of Gaussians with diagonal covariances.

Let $\gamma_t (i)$ denote the occupancy probability (or posterior probability) of $x_t$ being generated by the $i$-th component of GMM:
\begin{equation}
\gamma_t (i) = p(i | x_t) = \frac{w_i p_i (x_t | \lambda)}{\sum_{j = 1}^N w_j p_j (x_t | \lambda)}.
\end{equation}
Letting the subscript $d$ denote the $d$-th dimension of a vector, Fisher scores corresponding to GMM parameters are obtained as
\begin{align}
\label{eq:gmm_fv_0}
\frac{\partial \mathcal{L} (X | \lambda)}{\partial w_i} &= \sum_{t=1}^{T} \left[ \frac{\gamma_t (i)}{w_i} - \frac{\gamma_t (1)}{w_1} \right] \; \mathrm{for} \; i \ge 2, \\
\frac{\partial \mathcal{L} (X | \lambda)}{\partial \mu_{id}} &= \sum_{t=1}^{T} \gamma_t (i) \left[ \frac{x_{td} - \mu_{id}}{\sigma_{id}^2} \right], \\
\frac{\partial \mathcal{L} (X | \lambda)}{\partial \sigma_{id}} &= \sum_{t=1}^{T} \gamma_t (i) \left[ \frac{(x_{td} - \mu_{id})^2}{\sigma_{id}^3} - \frac{1}{\sigma_{id}} \right].
\end{align}
The gradient vector $G^{X}_{\lambda}$ in Eq.~(\ref{eq:1}) is obtained by concatenating these partial derivatives.

Next, the normalization terms, the Fisher information matrix $F_{\lambda}$ in Eq.~(\ref{eq:fim}) should be computed.
Let $f_{w_i}$, $f_{\mu_{id}}$, and $f_{\sigma_{id}}$ denote the terms on the diagonal of $F_{\lambda}$ which correspond to ${\mathcal{L} (X | \lambda)}/{\partial w_i}$, ${\partial \mathcal{L} (X | \lambda)}/{\partial \mu_{id}}$, and ${\partial \mathcal{L} (X | \lambda)}/{\partial \sigma_{id}}$ respectively.
In~\cite{per_cvpr07}, these terms are obtained approximately as
\begin{align}
f_{w_i} &= T \left( \frac{1}{w_i} + \frac{1}{w_1} \right), \\
f_{\mu_{id}} &= \frac{T w_i}{\sigma_{id}^2}, \\
f_{\sigma_{id}} &= \frac{2 T w_i}{\sigma_{id}^2}.
\end{align}
The gradient vector in Eq.~(\ref{eq:gmm_fv_0}) is related to the BoVW because the BoVW can be considered as the relative numbers of occurrences of words given by $\frac{1}{T} \sum_{t=1}^T \gamma_t (i) \; (1 \le i \le N)$.
While the BoVW captures 0-th order statistics, the Fisher kernel also captures 1-st and 2nd order statistics, resulting $(2 D + 1) N - 1$ dimensional vector.
The gradient vector corresponding to 0-th order statistics (${\mathcal{L} (X | \lambda)}/{\partial w_i}$) is sometimes not used because it does not contribute to performance \cite{per_cvpr07}.
In this case, the dimensionality of the GMM Fisher vector becomes $2 N D$.

\subsubsection{Improved Fisher Vector}
Although the above Fisher vector has achieved moderate performance, the advantage of this approach is considered to be its efficiency: it can create disctiminative high dimensional vector with small vocabularies (codebook size).
However, after the improved Fisher vector is proposed in \cite{per_eccv10}, it becomes widely used in both image classification~\cite{san_ijcv13} and image retrieval problems~\cite{per_cvpr10, jeg_pami12}.
The improved Fisher vector is calculated by applying two normalizations: power-normalization and $\ell_2$ normalization.
Power-normalization is to apply the following function to each of dimensions of the original Fisher vector:
\begin{equation}
f(z) = \mathrm{sign}(z) |z|^\alpha.
\end{equation}
The value of $\alpha = 0.5$ is often used for reasonable improvement.

\subsubsection{Other Extensions}
The Fisher vectors of the other probabilistic model is also proposed.
In \cite{kle_cvpr15}, the Fisher vectors of Laplacian Mixture Model (LMM) and a Hybrid Gaussian-Laplacian Mixture Model (HGLMM) are proposed.
In \cite{uch_acpr13}, the Fisher vectors of Bernoulli Mixture Model (BMM) is proposed for local binary features.
The Fisher vector is also improved by being combined with recent deep learning architechtures in image classification problems~\cite{sim_nips13, per_cvpr15} and image retrieval problems~\cite{cha_sp16, mor_arx16}.

\subsection{Vector of Locally Aggregated Descriptors}
\label{sec:vlad}
In \cite{jeg_cvpr10}, J{\'e}gou et al. have proposed an efficient way of aggregating local features into a vector of fixed dimension, namely Vector of Locally Aggregated Descriptors (VLAD).
In the construction of VLAD, VWs $c_1, \cdots, c_i, \cdots, c_N$ are first created by the $k$-means algorithm in the same way as in the BoVW framework.
Then, each feature vector $x$ is assigned to the closest VW $c_i$ ($i = \mathrm{NN}(x)$) in the visual codebook, where $\mathrm{NN}(x)$ denotes the identifier of VW closest to $x$.
For each of the visual words, the residual $x - c_i$ from assigned feature vector $x$ is accumulated, and the sums of residuals are concatenated into a single vector, VLAD.
More precisely, the VLAD vector $v$ is defined as
\begin{equation}
\label{eq:vlad_def}
v_{ij} = \sum_{x \; s.t. \; \mathrm{NN}(x) = i} \left[ x_j - c_{ij} \right],
\end{equation}
where $x_j$ and $c_{ij}$ denote the $j$-th component of the feature vector $x$ and the $i$-th VW, respectively.
Finally, the VLAD vector is $\ell_2$-normalized as $v := v / ||v||_2$.
VLAD can be considered as the simplified non-probabilistic version of the partial GMM Fisher vector corresponding to only the parameter $\mu_{id}$~\cite{jeg_pami12}.
Although the performance of VLAD is about the same or a little worse than the Fisher vector~\cite{jeg_pami12}, the VLAD has been widely used in image retrieval due to its simplicity.
There is many literature which extends the original VLAD~\cite{jeg_cvpr10}.
In the following, these extensions are briefly reviewed.

\subsubsection{Modified Normalizations}
Many literature focuses on the normalization step in order to improve the VLAD representation.
In \cite{jeg_pami12}, power-normalization is introduced as for the Fisher vector:
\begin{equation}
v_{ij} := \mathrm{sign}(v_{ij}) |v_{ij}|^\alpha,
\end{equation}
with $0 \le \alpha \le 1$.
This power-normalization is followed by $\ell_2$ normalization.
It have been shown that power-normalization consistently improves the quality of the VLAD representation~\cite{jeg_pami12}.
One interpretation of this improvement is that it reduces the negative influence of bursty visual elements~\cite{jeg_cvpr09}.
Regarding the parameter $\alpha$, $\alpha = 0.5$ is often used because it empirically shown to lead to near-optimal results.
Therefore, power-normalization is also referred to as Signed Square Root (SSR) normalization~\cite{jeg_eccv12, ara_cvpr13}.

In \cite{ara_cvpr13}, the other normalization is proposed, called intra-normalization.
In intra-normalization, the sum of residuals is independently $\ell_2$ normalized within each VLAD block $v_i$:
\begin{equation}
v_{ij} := v_{ij} / ||v_i||_2.
\end{equation}
Intra-normalization is also followed by $\ell_2$ normalization.
It is claimed that this normalization completely suppresses the burstiness effect regardless of the amount of bursty elements while power-normalization only discounts the burstiness effect~\cite{ara_cvpr13}.
After power-normalization, the standard deviations of the VLAD vectors become similar among all dimensions.
In other words, all dimensions can equally contribute to image similarity, improving the performance of the VLAD representation.

In \cite{del_mm13}, residual-normalization is proposed, where the residuals are normalized before summation so that all feature vectors contribute equally. With residual-normalization, Eq.~(\ref{eq:vlad_def}) is modified to
\begin{equation}
\label{eq:resi_norm}
v_{ij} = \sum_{x \; s.t. \; \mathrm{NN}(x) = i} \left[
\frac{x_j - c_{ij}}{||x - c_{i}||_2}
\right].
\end{equation}
It is shown that residual-normalization improves the performance of the VLAD representation if it is used in conjunction with power-normalization while it does \textit{not} without power-normalization~\cite{del_mm13}.
In \cite{jeg_cvpr14}, triangulation embedding is proposed.
It can be seen a modified version of VLAD, where normalized residuals from \textit{all} centroids are aggregated as
\begin{equation}
v_{ij} = \sum_x \left[
\frac{x_j - c_{ij}}{||x - c_{i}||_2}
\right].
\end{equation}
It is similar to residual-normalization in Eq.~(\ref{eq:resi_norm}) but only  residuals from the nearest centroids are aggregared in Eq.~(\ref{eq:resi_norm}).

\subsubsection{Other Extentions}
In~\cite{che_sp13}, it is proposed to modify the summation term in Eq.~(\ref{eq:vlad_def}) to mean or median operations.
It is claimed that the mean aggregation outperforms the original sum aggregation in terms of an image-level Receiver Operating Characteristic (ROC) curve analysis.
However, in \cite{spy_tmm14}, it is shown that the original sum aggregation is still better in terms of image retrieval performance.

In~\cite{jeg_eccv12, jeg_pami12}, it is proposed to perform Principal Component Analysis (PCA) to input vector $x$ before aggregation, which decorrelates and whiten input vector.
Decorrelating the input vector $x$ is very important because the VLAD implicitly assumes that the covariance matrices of $x$ is isotropic.
In~\cite{del_mm13}, Local Coordinate System (LCS) is proposed, where the residuals to be summed are rotated by a rotation matrix $Q_i$
\begin{equation}
v_{ij} = \sum_{x \; s.t. \; \mathrm{NN}(x) = i} Q_i \left[ x_j - c_{ij} \right].
\end{equation}
The VW-specific rotation matrix $Q_i$ is obtained by learning a local PCA per VW.
This is a contrast to the approach in \cite{jeg_pami12}, where the input vector $x$ is rotated by a globally learnt PCA matrix.
LCS has no effect if power-normalization is not applied ($\alpha = 1$).
However, it is shown that LCS with power-normalization outperforms the global rotation with power-normalization.

\section{Deep Learning for Image Retrieval}
Starting from ImageNet Large-scale Visual Recognition Challenge (ILSVRC) in 2012\footnote{\url{http://www.image-net.org/challenges/LSVRC/2012/}}, where Convolutional Neural Networks (CNN)~\cite{kri_nips12} had beaten the traditional state-of-the-art framework (i.e. SIFT feature + Fisher vector), CNN has become a de facto standard for image recognition tasks.
Following this trend, the deep learning approach also began to be applied to image retrieval tasks.
In this section, we briefly review recent deep learning approaches related to image retrieval.

Early works that have applied deep learning to image retrieval can be found in \cite{bab_eccv14, raz_cvprw14, sha_arxiv14, wan_mm14, cha_sp16}.
In \cite{bab_eccv14}, the CNN architecture used in \cite{kri_nips12} is applied for image retrieval.
Different from image recognition tasks, the best performance is achieved at the layer that is two levels below the outputs, not the very top of the network, which is consistent with the results of subsequent papers\footnote{In many papers, it is reported that the best performance has been achieved at the first fully connected layer.}.
In \cite{bab_eccv14}, the performance of CNN in \cite{kri_nips12} is reported to  be comparable to the Fisher vector or VLAD methods so far.
In \cite{raz_cvprw14, sha_arxiv14, wan_mm14}, a comparative study of CNN and the Fisher vector is performed.
In \cite{cha_sp16}, manys best practices for CNN is presented.

While the above methods utilizes CNN to extract a single global feature, there are different approaches to achieve better performance.
In \cite{lon_nips14}, it is shown that CNN can be applied to keypoint prediction task and find correspondences between objects.
In \cite{gon_eccv14, cim_cvpr15}, CNN features are densely extracted and the Fisher vector or VLAD framework is used for pooling (aggregation).
In \cite{liu_icmr15}, the dense CNN features from multiple networks are indexed by traditional inverted index.
In \cite{xie_icmr15}, a unified framework for both of image retrieval and classification is proposed, where CNN features are extracted from multiple object proposals for each image, and the Naive-Bayes Nearest-Neighbor (NBNN) search~\cite{boi_cvpr08} is performed to calculate the distance between a query image and reference images.
In \cite{ng_cvprw15}, \textit{convolutional features} are extracted from every position of different layers on CNN, and then these features are encoded by the VLAD framework.
It is reported that this framework achieves the best performance in very low-dimensional representation.
While the above methods utilizes the Fisher vector or VLAD for aggregation, in \cite{bab_iccv15}, it is claimed that the simple aggregation method based on sum pooling provides the best performance for deep convolutional features.

Deep learning is also applied for patch-level tasks.
In \cite{pau_iccv15}, in order to extract patch-level descriptors, Mairal et al. proposed a deep convolutional architecture based on Convolutional Kernel Network (CKN)~\cite{mai_nips14}, which is an unsupervised framework to learn convolutional architectures.
In \cite{sim_iccv15}, a Siamese network is used to learn discriminant patch representations, where an aggressive mining strategy is adopted to handle \textit{hard} negative and \textit{hard} positive pairs.
In \cite{zag_cvpr15}, similarity between image patches is directly learnt with CNN.
While several types of network are proposed and compared, it is reported that a two-channel and two-stream architecture achieved the best performance, where two patches to be compared are fed to the first convolutional layer directly (cf. a Siamese network).
For each patch, two regions with different scales are used as an input of the network.
Similarly, in \cite{han_cvpr15}, a patch matching system called MatchNet is proposed to learn a CNN for local feature description as well as a network for robust feature comparison.
In \cite{ver_cvpr15}, a new regression-based approach is proposed to extract feature points that are especially robust repeatable under temporal changes.
In \cite{yi_cvpr16}, a learning scheme based on CNN is introduced to estimate a canonical orientation for local features.
Because it is difficult to explicitly define a \textit{correct} canonical orientation, it is proposed to  implicitly define a canonical orientation to be learnt such that minimizes the distances between descriptors of correct feature pairs.
In \cite{yi_arxiv16}, a DNN architecture is proposed that combines the three
components of standard pipelines for local feature matching, i.e detection, orientation assignment, and description, into a single differentiable network.

There are several approaches to compress CNNs in order to reduce memory requirements and/or speed up the recognition.
In \cite{gon_arxiv14}, it is proposed to utilize product quantization~\cite{jeg10} to compress CNN.
In \cite{kim_iclr16}, a low-rank matrix approximation is used to compress CNNs.
In \cite{ras_arxiv16}, it is proposed to binarize CNNs and input signals to compress CNNs and achieve faster convolutional operations.
In \cite{han_iclr16, han_isca16}, pruning, quantization, and huffman coding is applied to CNNs to achieve an energy-efficient engine.

{\small
\bibliographystyle{ieee}
\bibliography{refs}
}

\end{document}